\documentclass{article}

\usepackage[preprint]{neurips_2026}

\usepackage[utf8]{inputenc} %
\usepackage[T1]{fontenc}    %
\usepackage{hyperref}       %
\usepackage{url}            %
\usepackage{booktabs}       %
\usepackage{amsfonts}       %
\usepackage{nicefrac}       %
\usepackage{microtype}      %
\usepackage{xcolor}         %
\usepackage{algorithm}
\usepackage{algpseudocode}
\input{packages}

\makeatletter

\renewcommand{\@toptitlebar}{%
  \vskip 0.08in
}

\renewcommand{\@bottomtitlebar}{%
  \vskip 0.12in
}

\usepackage{titlesec}

\titleformat{\section}
  {\normalfont\large\bfseries\scshape\raggedright}
  {\thesection}
  {0.6em}
  {\MakeLowercase}

\titleformat{\subsection}
  {\normalfont\normalsize\itshape\raggedright}
  {\thesubsection}
  {0.6em}
  { }

\titleformat{\subsubsection}
  {\normalfont\normalsize\bfseries\raggedright}
  {\thesubsubsection}
  {0.6em}
  { }

\titlespacing*{\section}
  {0pt}{1.4ex plus 0.3ex minus 0.2ex}{0.7ex plus 0.2ex}

\titlespacing*{\subsection}
  {0pt}{1.1ex plus 0.3ex minus 0.2ex}{0.5ex plus 0.2ex}

\titlespacing*{\subsubsection}
  {0pt}{0.9ex plus 0.2ex minus 0.2ex}{0.4ex plus 0.2ex}

\makeatother

\title{Recipes for Steering and Scaling LLMs via Sampling}

\author{%
  Jiajun He$^1$\\
  \texttt{jh2383@cam.ac.uk}\And Zongyu Guo$^2$\\
  \texttt{zongyuguo@microsoft.com} \AND José Miguel Hernández-Lobato$^1$\\
  \texttt{jmh233@cam.ac.uk}\And Yuanqi Du$^{3,4}$\\
  \texttt{yuanqidu@microsoft.com}
}

\begin{document}

\maketitle

\begin{abstract}
Large Language Models (LLMs) are probabilistic models, typically defined by an autoregressive factorization. 
While recent work has begun to study richer target distributions beyond the base model, the sampling strategies remain highly inefficient.
In this paper, we present a flexible and theoretically grounded framework for steering and scaling autoregressive LLMs with sampling. 
Within this framework, we describe two algorithms---one based on Sequential Monte Carlo (SMC) and one based on Replica Exchange (RE)---that steer generation toward \emph{powering}, \emph{product} or  \emph{tilting} of the base model distribution. We illustrate this framework through scaling the generation quality of LLMs without external supervision or reward models. 
Experimental results demonstrate our methods scale more favorably than Best-of-N and standard MCMC baselines. Overall, this paper offers a systematic recipe for probabilistic inference with LLMs via sampling.

\end{abstract}

\section{Introduction}

Large language models (LLMs) have achieved remarkable success across language understanding, reasoning, code generation, and scientific problem solving \citep{wei2022chain,achiam2023gpt,romera2024mathematical,shojaee2024llm,du2025accelerating}. At their core, LLMs are probabilistic models: they define distributions over sequences through an autoregressive factorization, where each token is sampled conditionally on the previously generated prefix. 

Respecting the probabilistic nature of the LLMs, massive progress have been made to improve their inference process, from faster inference with speculative decoding to fine-tuned model inference with reinforcement learning and controlled decoding for targeted generation \citep{leviathan2023fast,rafailov2023direct,mudgal2023controlled}. In addition to asymptotically accurate inference algorithms, numerous methods have simply leveraged the stochasticity of LLMs to scale their inference accuracy with more compute, including evolutionary algorithm, Monte Carlo tree search, Best-of-N decoding, confidence guidance and more \citep{wang2022self,wangefficient,hao2023reasoning,fu2025deep}. 

Alternatively, the fundamental mechanism of LLMs has been studied from the probabilistic inference perspective. One notable example is the sharpening mechanism observed that fine-tuning progressively sharpens the base distribution from a pre-trained LLM \citep{huang2024self,wu2025invisible,liu2025understanding}. Recently, Markov chain Monte Carlo (MCMC) methods have also been developed to sample from the sharpened base distribution \citep{karan2025reasoning}. Furthermore, the coverage of the base LLM has also been increasingly studied and found to be crucial for the success of post-training procedure \citep{chen2025coverage}.

In this paper, we systematically study sampling targets and algorithms in LLM scaling and steering. We first introduce a class of flexible target distributions, including powered, product and tilted base distributions, to sample. Then, we formulate the autoregressive inference process as a time-dependent transport that bridges the initial empty sentence to the distribution of sampled complete sentences. We establish a generalized Bayes' rule behind LLMs. From there, we introduce principled sequential Monte Carlo and replica exchange algorithms for efficient inference. We demonstrate the flexibility of the framework by enhancing the capability of the base LLM by purely altering its inference process without any external reward on mathematical reasoning problems.

\section{Background}

\subsection{Autoregressive Large Language Models} 
\label{sec:bg_ar}

Let $\mathcal{V}$ be the vocabulary set and $\widetilde{\mathcal{V}}:= \mathcal{V} \cup \{\varnothing\}$. Define an extended state space $\mathcal{X}^N$ for variable-length sequences of length $l\leq N$ padded with empty tokens 
$\mathcal{X}^N := \left\{ x_{0:N} \in \widetilde{\mathcal{V}}^N: \exists\, l\in \{0, ..., N\}, \text{s.t.}\, x_{0:l} \in \mathcal{V}^l, x_{l+1:N} =\varnothing \right\}$. An autoregressive LLM parameterizes a factorized joint distribution over discrete tokens
\begin{align}
    p(x_{0:N}) = p(x_0)\prod_{i=1}^N p(x_i|x_{<i})
\end{align}

where $x_i \in \widetilde{\mathcal{V}}$ with the following absorbing constraint:
\begin{align}
p(x_i=\varnothing | x_{1:i-1}) = 1, \qquad \text{when} \;\;x_{i-1} = \varnothing
\end{align}
Sampling from LLMs is straightforward: one first draws $x_0\sim p(x_0)$, and then sequentially samples $x_i$ conditional on all previous prefix tokens until the empty token is sampled.

\subsection{Sampling From Intractable Target Distributions}
Assume we want to draw samples from some intractable target distribution \(\pi\).
\emph{What should we do?}

There are two commonly used strategies.
One strategy aims to find a tractable distribution \(q\) that covers \(\pi\). We then draw samples from \(q\) and select those that look plausible under \(\pi\). 
Another strategy aims to explore \(\pi\) step by step: starting from an initial guess, we repeatedly propose a small move in a new direction.
If the direction looks promising, we take the step; otherwise, we stay where we are.
Two widely used algorithms for these two strategies are importance sampling and Markov Chain Monte Carlo (MCMC), which we briefly explain below:

\paragraph{Importance Sampling}
To sample from \(\pi\), we first draw \(N\) samples from the tractable proposal distribution \(q\):
$
    x^{(1)},\ldots,x^{(N)} \sim q .$
Then, we calculate the \emph{importance weight}
\begin{align}
    w(x^{(n)}) = \frac{\pi(x^{(n)})}{q(x^{(n)})}.
\end{align}
Samples with larger weights are more plausible under \(\pi\), while samples with smaller weights are less useful.
We can therefore either reweight the samples or resample them according to the weights
\begin{align}
    \bar w(x^{(n)})
    =
    \frac{w(x^{(n)})}{\sum_{m=1}^N w(x^{(m)})}.
\end{align}
This gives an approximate weighted sample from the target distribution \(\pi\).
However, importance sampling can fail when \(q\) does not cover \(\pi\) well, because most samples may receive negligible weight.

\paragraph{Markov Chain Monte Carlo}
MCMC takes a different approach.
Instead of drawing independent samples from a proposal \(q\), it constructs a Markov chain
\begin{align}
    x_0 \to x_1 \to \cdots \to x_N
\end{align}
whose stationary distribution is \(\pi\).
Starting from an initial point \(x_t\), we propose a local move \(x_t \to x'\) with the proposal conditional distribution with density $q(x'|x_t)$.
If the proposed point is more plausible under \(\pi\), we tend to accept it; otherwise, we may reject it and remain at \(x_t\).
For example, in the Metropolis-Hastings algorithm, the move is accepted with probability
\begin{align}
    a(x_t,x')
    =
    \min\left\{
    1,
    \frac{\pi(x')q(x_t|x')}{\pi(x_t)q(x'|x_t)}
    \right\}.
\end{align}
If accepted, we set \(x_{t+1}=x'\); otherwise, we set \(x_{t+1}=x_t\).
In this way, MCMC gradually explores the high-probability regions of \(\pi\).
Unlike importance sampling, it does not require a proposal \(q\) that globally covers the target, but it can mix slowly if moving between important regions of \(\pi\) is difficult.
For example, when the target distribution $\pi$ is multimodal, MCMC can become inefficient, as the Markov chain can get stuck in a local mode and cannot explore the entire support well.

\subsection{Accelerated Sampling with Intermediate Distributions} 
\label{sec:bg_sampling}

IS and MCMC give us two prototypes for how to draw samples from intractable distributions.
However, as we discussed, both approaches can face challenges when the target is multimodal or when the proposal is less calibrated toward the target.
\emph{In what way can we improve these approaches?}

The common idea is \emph{annealing}.
In other words, we choose some simple distribution to start with, and create a sequence of intermediate distributions, going from the simple distribution to the more complicated target distribution.
Building on this idea, IS and MCMC develop their annealing variants: sequential Monte Carlo and Replica Exchange.

Let $\pi_0$ be a tractable prior distribution, and  $\pi_N$ be our complicated target.
We introduce a sequence of intermediate distributions between them, given by the annealing path $\pi_i \propto \pi_0^{(1-i/N)} \pi_N^{(i/N)}$. 

\paragraph{Sequential Monte Carlo} 
Sequential Monte Carlo (SMC) approximates the sequence $\{\pi_i\}_{i=0}^N$ with a population of weighted particles. 
Given particles approximately distributed according to $\pi_{i-1}$, SMC propagates them with a forward kernel $x_i \sim F_i(\cdot \mid x_{i-1})$, and corrects the mismatch between the proposal and the next target in the annealing path using an incremental importance weight. 
Introducing a backward kernel, $B_{i-1}(x_{i-1}\mid x_i)$, the incremental weight is
\begin{align}
   G_i(x_{i-1},x_i)
= \frac{\pi_i(x_i)}{\pi_{i-1}(x_{i-1})}\frac{B_{i-1} (x_{i-1}\mid x_i)}{F_i(x_i\mid x_{i-1})}.
\end{align}

Thus the particle weight is updated by $w_i = w_{i-1}G_i(x_{i-1},x_i)$.
The forward kernel is used as a proposal for moving samples forward along the path, while the backward kernel specifies the reverse process used to derive weights that correct the samples.

\paragraph{Replica Exchange} 
Replica Exchange (RE) designs a Markov Chain on the joint space for the joint distribution
\begin{align}
    \pi_0 \times \pi_1 \times \cdots \times \pi_N,
\end{align}
with one replica per intermediate distribution $\pi_i$.
A local Markov kernel is used to update each replica under its own target $\pi_i$. 
In addition, neighboring replicas are occasionally swapped, known as communication steps.
For a proposed swap between levels $i$ and $i-1$, the Metropolis-Hastings acceptance probability is
\begin{align}
 A_{i, i-1}
    =
    \min\left\{1, 
    \frac{\pi_i(x_{i-1})}{\pi_i(x_i)}\frac{\pi_{i-1}(x_{i})}{\pi_{i-1}(x_{i-1})}\right\}.
\end{align}

\paragraph{Accelerated RE}
We can also introduce a forward $F$ and backward kernel $B$ for RE \citep{zhang2025accelerated}.
Starting from $x_i$ at the chain corresponding to $\pi_i$ and  $y_{i-1}$ at  $\pi_{i-1}$, we propose a ``swap" by
\begin{align}
    x_{i-1} \sim B_{i-1}(x_{i-1}|x_i), \quad   y_{i} \sim F_{i}(y_{i}|y_{i-1}).
\end{align}
Then, the acceptance rate is
\begin{align}
 A_{i, i-1}
    =
    \min\left\{1, 
    \frac{\pi_{i}(y_i)B_{i-1}(y_{i-1}|y_i)}{\pi_{i-1}(y_{i-1})F_i(y_i|y_{i-1})} \frac{\pi_{i-1}(x_{i-1})F_i(x_i|x_{i-1})}{\pi_{i}(x_i)B_{i-1}(x_{i-1}|x_i)} \right\}.
\end{align}
When kernels are chosen or learned well, this new variation of RE will have better performance (i.e., a higher acceptance rate between different distributions) than the standard version.

\section{Methods}

In this section, we establish our methodology. We first present the perspective of considering an LLM as a time-dependent transport. Next, we move to a flexible sampling target for LLM inference. Furthermore, we establish principled SMC and RE algorithms to sample from the flexible targets with practical implementation considerations.

\subsection{LLM as a Time-dependent Transport}

An LLM defines a forward Markov transition kernel $p(x_i|x_{i-1})$, as discussed in \Cref{sec:bg_ar}. We can then define a forward transition kernel for the transport
\begin{equation}
F_i(x_{0:i}\mid x_{0:i-1})
    =
    p(x_i\mid x_{<i}).
\end{equation}
We can also obtain a backward kernel from Bayes' rule:
\begin{equation}\label{eq:bayes_rule}
    B_{i-1}(x_{0:i-1}\mid x_{0:i})
    =
    \frac{
        p_{i-1}(x_{0:i-1})F_i(x_{0:i}\mid x_{0:i-1})
    }{
        p_i(x_{0:i})
    } = 1.
\end{equation}
Therefore, for the base autoregressive model, the backward transport is deterministic: it simply removes the last generated token. 
The nontrivial part of autoregressive generation lies in the forward kernel, which appends a token according to $p(x_i\mid x_{<i})$. 

After defining the transition kernel, we are able to run SMC or the accelerated RE to draw samples from some intractable distribution.
In the next section, we discuss the target distributions we consider.

\subsection{Sampling from Flexible Target Distributions}

\paragraph{Flexible target distributions} Inspired by recent works in controlling diffusion models \citep{wu2023practical,skreta2025feynman,singhal2025general,he2025rne}, we consider steering the distribution towards a new distribution $q(x_{0: N})$.
Some examples of this new distribution include (1) the  ``power"  version of original distribution $\pi(x_{0:N}) \propto p^\beta(x_{0:N})$, as considered by \citet{karan2025reasoning};
(2) the tilted distribution $\pi(x_{0:N}) \propto p(x_{0:N})\exp(r(x_{0:N}))$  with a reward, likelihood or verifier function $r$, which was also considered by \citet{zhao2024probabilistic};
(3) the product or quotient of distributions of two (or more) LLMs $\pi(x_{0:N}) \propto p(x_{0:N})p'(x_{0:N})$;
(4) any combination of the previous examples.
In summary, we consider the following task:

\begin{tcolorbox}[
    title={Examples of Steering Objectives},
    colback=gray!5,
    colframe=black!60,
    fonttitle=\bfseries,
    boxrule=0.3pt,
]
\small
We consider steering an auto-regressive model
$p(x_{0:N}) = p(x_0)\prod_{i=1}^N p(x_i \mid x_{<i})$
towards a target distribution $\pi(x_{0:N})$ of the following forms:

\begin{enumerate}[leftmargin=*]
    \item \textbf{Powered distribution}:
    \begin{align}
        \pi(x_{0:N}) \propto p(x_{0:N})^{\beta}
    \end{align}
    \item \textbf{Tilted (reward-conditioned) distribution}:
    \begin{align}
        \pi(x_{0:N}) \propto p(x_{0:N}) e^{r(x_{0:N})}
    \end{align}
    \item \textbf{Product/quotient of experts (multiple LLMs)}:
   \begin{align}
        \pi(x_{0:N}) \propto p(x_{0:N})\, p'(x_{0:N}) \text{ \ \ \ or \ \ \ } p(x_{0:N})/ p'(x_{0:N})
    \end{align}
    \item \textbf{Compositions} of the above objectives.
\end{enumerate}
\end{tcolorbox}
\par

Now, \emph{how can we obtain sample(s) $x_{0:N}\sim \pi(x_{0:N})$?}
This is a standard sampling task.
Similar to what was discussed by \citet{karan2025reasoning}, while the base LLM's distribution, $p(x_{0:N})=p(x_{0}) \prod_{i=1}^N p(x_{i}|x_{<i})$,  is factorized autoregressively, the corresponding target as defined above does not typically have a similar factorization structure.
Because of this, \citet{karan2025reasoning} introduced an MCMC algorithm to obtain samples from the power distribution.
However, as discussed in \Cref{sec:bg_sampling}, directly running MCMC can be less effective when the target distribution is complicated.
On the other hand, the LLM itself already defines a natural ``annealing" in the form of different resulting sequence lengths.
For example, we can define a sequence of annealing distributions $\pi(x_{0:i}) \propto p(x_{0:i})^\beta $ for power, $\pi(x_{0:i}) \propto p(x_{0:i}) e^{r(x_{0:i})} $ for tilting, etc.
Leveraging this property, we introduce two algorithms based on SMC and RE for more efficient sampling.

\paragraph{Sampling with Sequential Monte Carlo}

Following the particle reweighting for general SMC algorithms in \Cref{sec:bg_sampling}, we define the intermediate target distributions and we can obtain the corresponding incremental weights:
 
\textbf{Powered distribution}:
$
        \pi(x_{0:i}) \propto p(x_{0:i})^{\beta}, $ and $
        G(x_{0:i}, x_{0:i-1}) = \frac{p(x_{0:i})^\beta}{p(x_{0:i-1})^\beta}\frac{B_{i-1} (x_{0:i-1}\mid x_{0:i})}{F_i(x_{0:i}\mid x_{0:i-1})}.$
\textbf{Tilted (reward-conditioned) distribution}:
 $\pi(x_{0:i}) \propto p(x_{0:i}) e^{r(x_{0:i})}$, and $ 
        G(x_{0:i}, x_{0:i-1})
        =
        \frac{
            p(x_{0:i}) e^{r(x_{0:i})}
        }{
            p(x_{0:i-1}) e^{r(x_{0:i-1})}
        }
        \frac{
            B_{i-1}(x_{0:i-1}\mid x_{0:i})
        }{
            F_i(x_{0:i}\mid x_{0:i-1})
        }.$ 
\textbf{Product/quotient of experts (multiple LLMs)}:
 $
        \pi(x_{0:i}) \propto p(x_{0:i})\, p'(x_{0:i}), $ and $
        G(x_{0:i}, x_{0:i-1})
        =
        \frac{
            p(x_{0:i})p'(x_{0:i})
        }{
            p(x_{0:i-1})p'(x_{0:i-1})
        }
        \frac{
            B_{i-1}(x_{0:i-1}\mid x_{0:i})
        }{
            F_i(x_{0:i}\mid x_{0:i-1})
        }.$
Let's take a closer look at these distributions taking the power case as an example:
$G(x_{0:i}, x_{0:i-1}) = \frac{p(x_{0:i})^\beta}{p(x_{0:i-1})^\beta}\frac{B_{i-1} (x_{0:i-1}\mid x_{0:i})}{F_i(x_{0:i}\mid x_{0:i-1})}$.
This formula indicates that, every time we sample the next token, we need to evaluate the entire sequence likelihood $p(x_{0:i})$, which can be prohibitive in LLM sampling.

Fortunately, recall that the base distribution is factorized as an AR model, hence, we have
\begin{align}
    \frac{p(x_{0:i})^\beta}{p(x_{0:i-1})^\beta} = p(x_i|x_{<i})^\beta
\end{align}
Similarly, using also an LLM as the proposal, with \Cref{eq:bayes_rule}, we have
\begin{align}
 \frac{B_{i-1} (x_{0:i-1}\mid x_{0:i})}{F_i(x_{0:i}\mid x_{0:i-1})} = \frac{q_{i-1}(x_{0:i-1})}{q_{i}(x_{0:i})} = \frac{1}{q(x_i|x_{<i})}
\end{align}
Here we use $q$ to denote that the proposal LLM, which can be different from our target LLM.
This allows us to obtain the following SMC weights:
\begin{tcolorbox}[
    title={Sequential Monte Carlo Weights},
    colback=gray!5,
    colframe=black!60,
    fonttitle=\bfseries,
        boxrule=0.3pt,
]
\small
Assume we have a sequence $x_{0:i-1}$, and we propose with an proposal LLM $q$ by
 $x_{0:i} \sim  F_{i}(  x_{0:i}|   x_{0:i-1}).
$
The incremental SMC weight is given by
\begin{enumerate}[leftmargin=*]
    \item \textbf{Powered distribution}:
  \begin{align}
        \pi(x_{0:i}) \propto p(x_{0:i})^{\beta},\qquad
        G(x_{0:i}, x_{0:i-1}) = \frac{p(x_i|x_{<i})^\beta}{q(x_i|x_{<i})}
  \end{align}
    \item \textbf{Tilted (reward-conditioned) distribution}:
   \begin{align}
        \pi(x_{0:i}) \propto p(x_{0:i}) e^{r(x_{0:i})}, \quad
        G(x_{0:i}, x_{0:i-1})
        = \frac{p(x_i|x_{<i})e^{r(x_{0:i})} }{q(x_i|x_{<i})  e^{r(x_{0:i-1})}} .
   \end{align}
    \item \textbf{Product/quotient of experts (multiple LLMs)}:
   \begin{align}
        \pi(x_{0:i}) \propto p(x_{0:i})\, p'(x_{0:i}) , \quad
        G(x_{0:i}, x_{0:i-1})
        =
        \frac{p(x_i|x_{<i})p'(x_i|x_{<i})}{q(x_i|x_{<i})}.
    \end{align}
\end{enumerate} 
\end{tcolorbox}

\paragraph{Sampling with Replica Exchange}
Similarly, we can also derive the Replica Exchange rate with target base LLM $p$ and proposal LLM $q$, as stated below:

\begin{tcolorbox}[
    title={Replica Exchange Swap Ratios},
    colback=gray!5,
    colframe=black!60,
    fonttitle=\bfseries,
        boxrule=0.3pt,
]
\small
Assume we have sequences $x_{0:i-1}$ and $y_{0:i}$, and we propose to ``swap" them using the proposal LLM $q$:
\begin{align}
    y_{0:i-1} \sim  B_{i-1}(  y_{0:i-1}|   y_{0:i}), \quad  x_{0:i} \sim  F_{i}(  x_{0:i}|   x_{0:i-1}).
\end{align}
The acceptance rate is $A = \min \left\{1, \alpha \right\}$, with $\alpha$ defined according to the following cases:
\begin{enumerate}[leftmargin=*]
    \item \textbf{Powered distribution} $\pi  \propto p ^{\beta}$:
    \begin{align}
        \alpha
        =
        \frac{
            p(x_{i|<i})^{\beta} 
        }{
           p(y_{i|<i})^{\beta}   
        }\frac{q(y_{i|<i}) }{
         q(x_{i|<i})}.
    \end{align}
    \item \textbf{Tilted (reward-conditioned) distribution} $\pi 
        \propto 
        p  e^{ r }$:
    \begin{align}
         \alpha
        =
        \frac{
            e^{r(x_{0:i}) - r(x_{0:i-1})}
        }{
          e^{r(y_{0:i}) - r(y_{0:i-1})}
        }
         \frac{
            p(x_{i|<i}) 
        }{
           p(y_{i|<i})  
        }\frac{q(y_{i|<i}) }{
         q(x_{i|<i})}.
    \end{align}
    
    \item \textbf{Product of experts (multiple LLMs)} $\pi 
        \propto 
        p  p' $:
        \begin{align}
            \alpha =  \frac{
            p(x_{i|<i}) 
        }{
           p(y_{i|<i})  
        }\frac{
            p'(x_{i|<i}) 
        }{
           p'(y_{i|<i})  
        } \frac{q(y_{i|<i}) }{
         q(x_{i|<i})}.
        \end{align}
\end{enumerate}
\end{tcolorbox}

\emph{We also highlight that the RE algorithm we proposed is not a direct application of standard replica exchange but belongs to a new family of RE methods. }
In standard RE, one constructs a ladder of distributions by annealing a difficult target into simpler ones, with all replicas living in the same state space.
Recent path-space extensions, such as \citet{zhang2025accelerated}, lift RE from endpoint distributions to path measures, but still require defining a ladder of path distributions over a common path space.

By contrast, our construction applies RE directly to the autoregressive generation \emph{process}. 
The replicas evolve through chunks of tokens, and these replicas differ in length and hence are not in the same space. 
This gives a different and more flexible form of replica exchange, naturally suited to variable-length and autoregressive state spaces.\vspace{-10pt}

\paragraph{SMC and RE in Chunks}
Although we present the SMC weight and RE swap ratio for a single incremental step from $i-1$ to $i$, in practice, both algorithms can be applied in ``chunks", corresponding to a contiguous sequence of tokens.
Given a chunk size $B$, the algorithm moves between chunk states that differ by $B$ autoregressive steps (e.g., from $i-B$ to $i$).
The weight or swap ratio for the whole chunk is then obtained by multiplying the corresponding single-step quantities within that chunk.
We summarize the algorithm for SMC and RE in \Cref{fig:llm_chunked_sampling_algorithms}.\vspace{-10pt}
\begin{figure}[t]
\centering
\begin{minipage}[t]{0.49\linewidth}
\begin{algorithm}[H]
\caption{SMC}
\label{alg:llm_chunked_smc}
\begin{algorithmic}[1]
\Require proposal LLM \(q\), particles \(K\), length \(N\), chunk size \(B\)
\State Let \(n_m=mB\), with \(M=N/B\)
\State Initialize \(x_{0:n_0}^{(k)}\sim q(x_{0:n_0})\), \(w^{(k)}\gets 1\)
\For{\(m=1,\ldots,M\)}
    \For{\(k=1,\ldots,K\)}
    \State {\color{gray}\# extend each sample one chunk}
        \For{\(j=n_{m-1}+1,\ldots,n_m\)}
            \State Sample \(x_j^{(k)}\sim q(\cdot\mid x_{<j}^{(k)})\)
        \EndFor
        \State \(x_{0:n_m}^{(k)}=(x_{0:n_{m-1}}^{(k)},x_{n_{m-1}+1:n_m}^{(k)})\)
        \State 
       $            w^{(k)}
            \gets
            w^{(k)}
            G_m (x_{0:n_m}^{(k)},x_{0:n_{m-1}}^{(k)})
        $
    \EndFor
    \State Normalize \(\{w^{(k)}\}_{k=1}^K\)
        \State Resample particles 
        \State Reset \(w^{(k)}\gets 1/K\)
\EndFor
\State \Return \(\{x_{0:N}^{(k)},w^{(k)}\}_{k=1}^K\)
\end{algorithmic}
\end{algorithm}
\end{minipage}
\hfill
\begin{minipage}[t]{0.49\linewidth}
\begin{algorithm}[H]
\caption{RE}
\label{alg:llm_chunked_re}
\begin{algorithmic}[1]
\Require proposal LLM \(q\), iterations \(K\), length \(N\), chunk size \(B\)
\State Let \(n_m=mB\), with \(M=N/B\)
\State Initialize initial guess \(z_m\sim q(x_{0:n_m})\)
\For{\(s=1,\ldots,K\)}
    \For{\(m=1,3, 5\ldots\) or \(2,4, 6\ldots\)}
        \State Let \(x_{0:n_{m-1}}=z_{m-1}\), \(y_{0:n_m}=z_m\)
        \State {\color{gray}\# remove one chunk from $y_{0:n_{m-1}}$}
        \State Set
        $
            \tilde y_{0:n_{m-1}} = y_{0:n_{m-1}}
        $
        \State {\color{gray}\# extend one chunk from \(x_{0:n_{m-1}}\)}
        \For{\(j=n_{m-1}+1,\ldots,n_m\)}
            \State Sample \(\tilde x_j\sim q(\cdot\mid \tilde x_{<j})\)
        \EndFor
        \State Set \(\tilde x_{0:n_m}=(x_{0:n_{m-1}},\tilde x_{n_{m-1}+1:n_m})\)
        \State  
     $(z_{m-1},z_m)
            \leftarrow
            (\tilde y_{0:n_{m-1}},\tilde x_{0:n_m})$
        with probability
        $ 
            A_m=\min\{1,\alpha_m \}
        $
    \EndFor
\EndFor
\State \Return final-level replica \(z_M\)
\end{algorithmic}
\end{algorithm}
\end{minipage}
\caption{Sampling algorithms for flexible LLM inference targets. Here \(G_m \) is the chunk-level incremental SMC weight from length \(n_{m-1}\) to \(n_m\), and \(\alpha_m \) is the chunk-level Metropolis-Hastings ratio. 
For RE, swaps are performed in alternating adjacent pairs: when $s$ is odd, we swap between $(1, 2), (3, 4), ...$, and when $s$ is even, we swap between $(2, 3), (4, 5), ...$. Therefore, the budget for RE and SMC is the same when the RE iteration is twice the number of SMC particles. }
\label{fig:llm_chunked_sampling_algorithms}\vspace{-20pt}
\end{figure}

\paragraph{EOS Tokens}
Finally, we note that LLM generations are typically variable-length. 
For SMC, EOS tokens can be handled straightforwardly: once a particle reaches EOS, we stop extending that particle and keep it in the final answer pool, while the remaining particles continue to evolve.

For RE, we maintain the convention that each complete sequence ends with an EOS token. During initialization (initial guess), we generate until reaching EOS token, and divide them into chuncks, corresponding to each intermediate target.
During sampling, if a newly proposed chunk contains an EOS token and is accepted, we truncate the sequence at that EOS token and discard all subsequent chunks. Conversely, if the current final chunk does not contain an EOS token, we keep extending the sequence with additional chunks until an EOS token is reached.

\vspace{-5pt}
\section{Related Works}
\vspace{-5pt}

\subsection{Steering LLMs at Inference-time}

Since the emergence of general-purpose LLMs, a substantial body of work has studied how to elicit, control, and improve their behavior at inference time. Early methods focused on prompting, leveraging the conversational interface to induce intermediate reasoning or structured decompositions~\citep{kojima2022large,wei2022chain}. \citet{yang2023large} proposed to further optimize the text prompt to control the text output. In parallel, mechanistic and representation-centric approaches aimed for more direct control by mapping and intervening the internal representations of LLMs \citep{li2023inference,cunningham2023sparse,kong2024aligning}. Another branch of work integrates LLM inference with implicit or explicit search and external feedback \citep{valmeekam2022large,madaan2023self}. Among them, evolutionary algorithms demonstrated strong performances in searching mathematical programs, molecular structures and symbolic equations \citep{romera2024mathematical,novikov2025alphaevolve,wangefficient,shojaee2024llm}. The most relevant work to our setting is sampling-based approaches:\citet{mudgal2023controlled} approximated the optimal value function to control the sampling process; 
\citet{zhao2024probabilistic} applied twisted sequential Monte Carlo to steer LLMs; 
recently, \citet{karan2025reasoning} proposed to enhance LLM reasoning via sampling from a tempered distribution  (power) using MCMC.

\subsection{Inference-time Control with Sampling in Diffusion Models}

Inference-time control has also been extensively studied in other generative models, especially diffusion models. A common strategy is to modify the generation process using guidance so that samples satisfy additional constraints or exhibit desired properties~\citep{dhariwal2021diffusion,ho2022classifier,chung2022diffusion,kong2024diffusion}. 
While such heuristic guidance methods are often effective, they can introduce bias and may fail under strong constraints. 
This has motivated more principled sampling-based approaches, including annealed importance sampling, sequential Monte Carlo, Feynman-Kac-based correction, and replica-exchange-style methods~\citep{wu2023practical,skreta2025feynman,singhal2025general,du2023reduce,he2025crepe}. 
These methods use annealing intermediate distributions to progressively steer generation. Our work brings this sampling-based test-time control perspective to autoregressive LLM generation, where the generation trajectory is discrete, variable-length, and naturally factorized token by token.

\section{Experiments}

We evaluate whether the proposed sampling algorithms can improve LLMs at inference time. \vspace{-5pt}

\paragraph{Benchmarks and models.}
Following \citet{karan2025reasoning}, we evaluate on MATH500 \citep{lightman2023lets} and GPQA-Diamond \citep{rein2023gpqa}.
We use the same data split and answer format as \citet{karan2025reasoning}.
For MATH500, we evaluate both Qwen2.5-Math-7B and Qwen2.5-7B.
For GPQA-Diamond, we evaluate Qwen2.5-32B-Instruct, since we found that the 7B base model used in \citet{karan2025reasoning} does not reliably follow the required answer format, leading to unreliable correctness estimates.
All generations use chain-of-thought prompting.
For evaluation, we parse the last boxed expression from each generated solution and compare it against the ground-truth answer.\vspace{-5pt}

\paragraph{Sampling setup.}
Unless otherwise stated, we sample with maximum generation length \(3072\), proposal temperature \(0.25\), and \(16\) chunks.
Thus, each chunk contains at most \(192\) newly generated tokens.
The proposal distribution is the low-temperature autoregressive sampler, while the target distribution is a powered and optionally entropy-tilted version of the base model distribution.
In particular, the power parameter is \(\alpha=1/0.25=4\), and we additionally consider an intrinsic ``confidence" tilt  based on the model entropy.

Note that we do not tilt with the likelihood as the confidence, as this would lead to the same formulation as powering.
Instead, we use the model's predictive entropy as an intrinsic confidence signal.
More precisely, for a generated chunk \(x_{a:b}\), we tilt with \vspace{-5pt}
\begin{align}
    \log \gamma(x_{a:b})
    =
    -\lambda \bar H(x_{a:b}).
\end{align}
where $\bar H(x_{a:b})$ is  the average next-token entropy over the chunk:\vspace{-5pt}
\begin{align}
    \bar H(x_{a:b})=
    \frac{1}{b-a+1}
    \sum_{j=a}^{b} H_j , \text{ where } H_j=
    -\sum_{v\in\mathcal V}
    p(v|x_{<j})
    \log p(v|x_{<j}).
\end{align}
Equivalently, this upweights chunks where the base model has lower predictive entropy, i.e., higher confidence.
We sweep \(\lambda\in\{0,100,500,1000\}\).
Importantly, this tilt uses only the base model's own predictive uncertainty and does not rely on an external verifier, similar to \citep{fu2025deep}.

\paragraph{Research questions.} In the following, we consider five questions:
(1) do SMC and RE improve over the autoregressive base sampler and naive low-temperature sampling?
(2) do they provide better inference-time scaling than the MCMC sampler of \citet{karan2025reasoning}?
(3) can the model performance be improved without any external verifier using both powering and tilting?
(4) are these sampling methods better than simple Best-of-$N$?
and (5) since SMC and RE produce a population of samples rather than only a single selected output, can their sample diversity bring us more benefit?

\begin{figure}[t]
    \centering
    \begin{minipage}{0.46\linewidth}
        \centering
        \includegraphics[width=\linewidth]{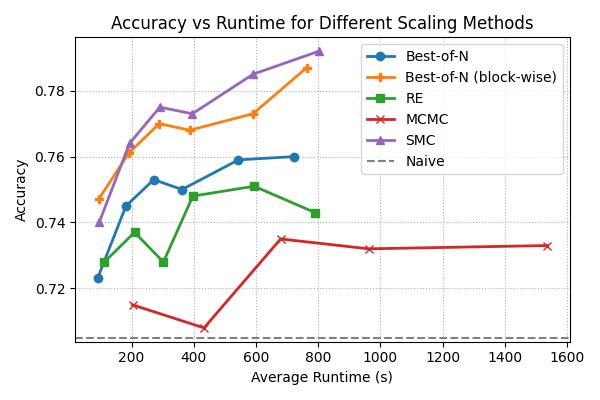}
        \caption{SMC, RE, MCMC, BON, BON with block, Naive Tempering on MATH500 with Qwen-2.5-Math-7B.}
        \label{fig:results_math}
    \end{minipage}
    \hfill\hfill
    \begin{minipage}{0.46\linewidth}
        \centering
        \includegraphics[width=\linewidth]{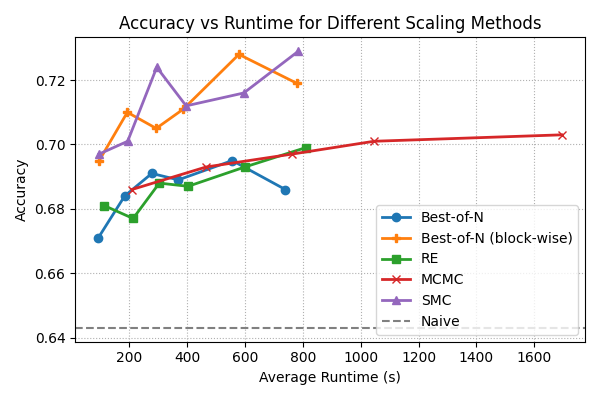}
        \caption{SMC, RE, MCMC, BON, BON with block, Naive Tempering on MATH500 with Qwen-2.5-7B.}
        \label{fig:results_base}\vspace{-5pt}
    \end{minipage}
\end{figure}

\paragraph{SMC and RE improve over naive autoregressive sampling.}
As shown in \Cref{fig:results_base,fig:results_math}, both SMC and RE substantially outperform naive low-temperature sampling on MATH500.
This follows the same argument as \citep{karan2025reasoning}, where both SMC and RE can be viewed as a global ``search", while the standard low-temperature sampling is greedy.

\paragraph{SMC and RE scale more favorably than MCMC.}
We then compare our proposed framework against the recent MCMC-based sampling method \citep{karan2025reasoning}.
More precisely, they consider powering the LLM output with MCMC.
From \Cref{fig:results_base,fig:results_math}, we can see that SMC and RE all present better scaling properties compared to MCMC.
The reason is that MCMC revises the trajectory by randomly removing and refilling.
This causes a dilemma:
if the sentence needs to be largely revised, we expect to refill a larger chunk.
However, when the chunk need to refill is larger, the acceptance rate typically drops drastically.
On the other hand, both SMC and RE are based on \emph{annealing} through a sequence of intermediate targets.
These targets serve as ``checkpoints", making the proposal being used more efficient.

\paragraph{Powering and entropy tilting improve performance without an external verifier.}
Beyond power, we now verify if the proposed entropy tilting could also improve the performance without an extra verifier.
In \Cref{tab:main_results}, we investigate this on both MATH500 and GPQA-diamond datasets.
We also sweep with different power and different tilting strengths.
For both SMC and RE, both of powering and tilting improve the accuracy of the base model, and combining them further improves the results.
This suggests that the model's own predictive entropy is a useful intrinsic confidence signal.

At the same time, the effect of entropy tilting is not monotone.
Very large tilt strengths can slightly hurt performance.
A possible reason is that a strong tilt makes the target and proposal too different, which decreases the acceptance rate, especially for RE.
This is also consistent with the observation in \Cref{tab:math500_entropy_tilting}, where RE tends to drop for larger strength while SMC tends to plateau.

\paragraph{Best-of-\(N\) is still a strong baseline.}
We also compare the sampling method with BON. 
We consider two different BON setups: for the simplest BON, we directly finish the generation and then select the best outcome according to the likelihood; for the second case, which we refer to as BON block-wise, we use the same chunk idea as SMC and RE. We select the best candidate after finishing each chunk, and then use this as the prefix for the next chunk.
The results in \Cref{fig:results_base,fig:results_math} show that Best-of-\(N\) is a strong baseline.
It not only outperforms MCMC, but also outperforms RE and on par with SMC.
However, we note that SMC and RE aim to draw samples from the entire distribution, while BON is searching for the best one.
Therefore, when we care about sample diversity, SMC or RE can be effective yet simple choices, while BON typically returns a single sample.

\begin{figure}[t]
    \centering
    \begin{subfigure}{0.82\linewidth}
        \centering
        \includegraphics[width=\linewidth]{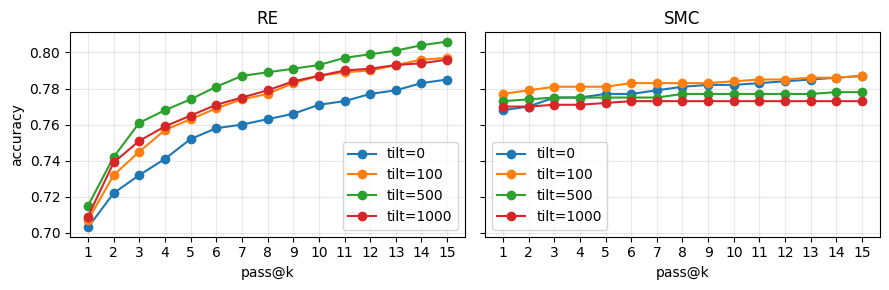}
        \caption{Qwen-7B-Math}
    \end{subfigure}
    \begin{subfigure}{0.82\linewidth}
        \centering
        \includegraphics[width=\linewidth]{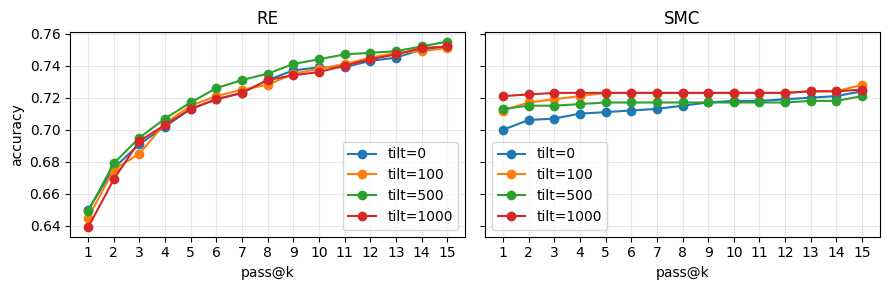}
        \caption{Qwen-7B}
    \end{subfigure}
    \caption{Pass@K results for SMC and RE. }
    \label{fig:passk}\vspace{-20pt}
\end{figure}

\paragraph{Population-based sampling gives additional pass@\(k\) benefits.}
Following the discussion above, we next ask whether the sample population produced by SMC and RE is useful beyond selecting a single final answer.
To evaluate this, we report pass@\(k\) accuracy in \Cref{fig:passk}.
For a fair comparison, we run SMC with \(15\) particles and for RE, we run \(30\) swap iterations.
This gives the same sampling budget to both methods, and both correspond to at most \(15\) final-level candidate sequences.
We then evaluate whether the correct answer appears among the \(k\) generated samples.
For RE, we simply take the first $k$ samples for evaluation, while for SMC, we randomly select $k$ from the final answer pool.

The pass@\(k\) curves show that both SMC and RE benefit from having multiple samples.
This confirms that the generated populations are not merely redundant copies of a single solution but contain useful alternative reasoning traces.
However, the two methods behave differently.
SMC gives stronger first-sample performance,  but at the same time, the pass@\(k\) curve tends to improve only mildly as \(k\) increases, reflecting that the samples are mostly collapsed.
RE exhibits the opposite behavior.
Although its first sample can be less competitive, its pass@\(k\) accuracy continues to increase more substantially with \(k\) and in the end outperform SMC.
This suggests that the RE population preserves more diverse candidates, which also aligns with the findings by recent work in diffusion control \citep{he2025crepe}.
Thus, SMC is especially effective when one wants a single strong answer, while RE is particularly useful when multiple candidate solutions can be exploited.

\begin{table}[t]
\centering

\begin{subtable}[t]{0.52\linewidth}
\centering
\resizebox{\linewidth}{!}{%
\begin{tabular}{llcccc}
\toprule
\textbf{Model} & \textbf{Method} & \textbf{0} & \textbf{100} & \textbf{500} & \textbf{1000} \\
\midrule
\multirow{3}{*}{Mathmodel} 
& RE & 0.746 & 0.756 & 0.774 & 0.769 \\
& SMC & 0.770 & 0.777 & 0.772 & 0.771 \\
& Naive Temp & \multicolumn{4}{c}{0.685} \\
\midrule
\multirow{3}{*}{Basemodel} 
& RE & 0.679 & 0.704 & 0.701 & 0.693 \\
& SMC & 0.702 & 0.711 & 0.717 & 0.720 \\
& Naive Temp & \multicolumn{4}{c}{0.633} \\
\bottomrule
\end{tabular}
}
\caption{MATH500 accuracy under different  tilting strengths.}
\label{tab:math500_entropy_tilting}
\end{subtable}
\hfill
\begin{subtable}[t]{0.47\linewidth}
\centering
\renewcommand{\arraystretch}{1.1}
\resizebox{\linewidth}{!}{%
\begin{tabular}{c|cc|cc}
\toprule
& \multicolumn{2}{c|}{\textbf{Tilt strength = 0}} 
& \multicolumn{2}{c}{\textbf{Tilt strength = 500}} \\
\textbf{Power} & \textbf{RE} & \textbf{SMC} & \textbf{RE} & \textbf{SMC} \\
\midrule
0.25 & 0.464 & 0.486 & 0.497 & 0.464 \\
0.50 & 0.491 & 0.458 & 0.503 & 0.505 \\
0.75 & 0.484 & 0.473 & \textbf{0.512} & 0.502 \\
1.00 & 0.483 & 0.483 & 0.497 & 0.477 \\
\bottomrule
\end{tabular}
}
\caption{GPQA results on Qwen2.5-Instruct-32B under different power values and tilt strengths.}
\label{tab:gpqa_re_smc_power_tilt}
\end{subtable}

\caption{Results on MATH500 and GPQA. We run both SMC and RE for the same budget (15 samples in SMC, or 30 iterations in RE).}
\label{tab:main_results}\vspace{-15pt}
\end{table}

\vspace{-6pt}
\section{Conclusion}
\vspace{-6pt}
We presented a probabilistic sampling framework for steering and scaling autoregressive LLMs at inference time.
By viewing autoregressive generation as a time-dependent transport, we derived flexible target distributions, including powered, tilted, and product-form objectives, together with practical SMC and RE algorithms for sampling from them.
Empirically, we showed that these methods improve over naive low-temperature sampling and scale more favorably than standard MCMC on reasoning benchmarks.
We also found that intrinsic entropy tilting can further improve performance without using an external verifier or reward model.

Our results highlight complementary strengths of the two samplers.
SMC is particularly effective for obtaining a strong single answer, while RE better preserves sample diversity and gives stronger pass@\(k\) gains.
Although Best-of-\(N\) remains a strong search baseline, SMC and RE provide a principled way to sample from the full target distribution rather than only selecting one high-scoring output.
We hope this work serves as a step toward a broader view of LLM as probabilistic inference over autoregressive trajectories and can inspire future work in LLM scaling and sampling.
\vspace{-4pt}

\paragraph{Limitations.}
Our method improves inference-time scaling through better sampling, but it also introduces additional computational overhead compared with standard decoding.
Its effectiveness depends on the quality of the underlying model, and a poor base model may get saturated quickly.
Additionally, in most tasks of LLMs, the exact form of the target distribution may not be important.
In such cases, as we have demonstrated, BON might be a simple alternative with strong performance.

\section*{Acknowledgment}
JH acknowledges support from the University of Cambridge Harding Distinguished Postgraduate Scholar Programme.
JMHL acknowledges funding from AI Hub in Generative Models, under grant EP/Y028805/1. YD acknowledges support from Cornell University.

\bibliographystyle{abbrvnat}
\bibliography{example_paper}

\appendix
\newpage

\section{Experimental Hyperparameters}
\label{app:hyperparameters}

We use the same MATH500 and GPQA-Diamond datasets as \citet{karan2025reasoning}.
For MATH500, we evaluate Qwen2.5-Math-7B and Qwen2.5-7B.
For GPQA-Diamond, we evaluate Qwen2.5-32B-Instruct.
All experiments use chain-of-thought prompting.

We use maximum generation length \(T_{\max}=3072\), proposal temperature \(\tau=0.25\), and \(16\) chunks.
Thus, each chunk contains at most \(3072/16=192\) newly generated tokens.

\section{Broader Impact}\label{broader_impact}
This paper studies inference-time scaling for LLMs through improved sampling methods.
Like many techniques that increase the capability of LLMs, our work may have broader societal implications.
However, these implications are not specific to our method but are instead associated with the development and deployment of LLMs more generally.
We advocate for the responsible and appropriate use of LLMs, with careful consideration of potential risks and impacts.

\section{Declaration  of Experiments compute resources}\label{resource}

We run our results on NVIDIA A100 40G GPUs and  NVIDIA A100 80G GPUs. 
Each question is fast, while collecting the results for all datasets and all methods can take several days to weeks.

\section{Declaration of LLM usage}\label{sec:llm_usage}
We used LLMs to assist with coding, manuscript polishing, and proofreading.
We also study relatively lightweight LLMs (Qwen-7B/32B) in this paper.

\section{Licenses for existing assets}\label{licence}
We use the mode weight of Qwen-2.5, released under the Qwen LICENSE AGREEMENT.
We use the code from by \citet{karan2025reasoning} at \url{https://github.com/aakaran/reasoning-with-sampling}, released without license.

GPQA and MATH500 are both released with MIT license.

\end{document}